\documentclass[a4paper,conference,final]{IEEEtran}
\IEEEoverridecommandlockouts

\usepackage[binary-units=true]{siunitx}
\DeclareSIUnit \dB {dB}
\DeclareSIUnit \dBm {dBm}
\DeclareSIUnit \dBi {dBi}

\usepackage{cite}
\usepackage{amsmath,amssymb,amsfonts}
\usepackage{algorithmic}
\usepackage{graphicx}
\usepackage{textcomp}
\usepackage{xcolor}
\usepackage{caption}
\usepackage{subcaption}

\usepackage{dsfont}
\usepackage{multirow}

\newcommand{\eqr}[1]{$#1$}

\begin{document}

\title{Location Anomalies Detection for\\ Connected and Autonomous Vehicles
}

\author{\IEEEauthorblockN{Xiaoyang Wang\IEEEauthorrefmark{1},
Ioannis Mavromatis\IEEEauthorrefmark{1}, Andrea Tassi\IEEEauthorrefmark{1}, Ra\'ul Santos-Rodr\'iguez\IEEEauthorrefmark{2} and Robert J. Piechocki\IEEEauthorrefmark{1}\IEEEauthorrefmark{3}}
\IEEEauthorblockA{\IEEEauthorrefmark{1} Department of Electrical and Electronic Engineering, University of Bristol, UK\\
\IEEEauthorrefmark{2} Department of Engineering Mathematics, University of Bristol, UK\\
\IEEEauthorblockA{\IEEEauthorrefmark{3}The Alan Turing Institute, London, NW1 2DB, UK}\\
Email: \{xiaoyang.wang, ioan.mavromatis, a.tassi, enrsr,  r.j.piechocki\}@bristol.ac.uk}}

\maketitle

\begin{abstract}


Future Connected and Automated Vehicles (CAV), and more generally ITS, will form a highly interconnected system. Such a paradigm is referred to as the Internet of Vehicles (herein Internet of CAVs) and is a prerequisite to orchestrate traffic flows in cities. For optimal decision making and supervision, traffic centres will have access to suitably anonymized CAV mobility information. Safe and secure operations will then be contingent on early detection of anomalies. In this paper, a novel unsupervised learning model based on deep autoencoder is proposed to detect the self-reported location anomaly in CAVs, using vehicle locations and the Received Signal Strength Indicator (RSSI) as features. Quantitative experiments on simulation datasets show that the proposed approach is effective and robust in detecting self-reported location anomalies.

\end{abstract}


\section{Introduction}

The past decades have witnessed rapid developments in Intelligent Transportation Systems (ITS). Connected and Autonomous Vehicles (CAV) are an integral part of ITS and will redefine mobility, change the existing vehicle usage and pave the way for future transportation services. Vehicles collect the necessary information via a number of onboard sensors. This information is later disseminated to the surrounding environment in a Vehicle-to-Everything (V2X) fashion, negotiating manoeuvres and building a more agile, safe and efficient traffic network~\cite{shladover2018connected}.

Without constant human supervision, the safety of CAVs heavily relies on the knowledge acquired from the connected surrounding environment~\cite{kockelman2016implications}.
However, this dependence of exchanged data brings into the surface several security risks and potential malicious cyberattack. The connected nature of the vehicles increases the risk of compromised vehicles on the road, and thus the demand for more sophisticated anomaly detection and cybersecurity protection techniques. In this work, we will focus on CAVs that maliciously exchange their falsified self-reported locations to the surrounding vehicles, presenting a novel way of detecting and counteracting on the abnormalities.


Anomaly detection is the identification of abnormal observations that do not conform to the expected behaviour. 
Ultimately, the goal is to present a quick and reliable alert when an anomaly occurs, helping the system to respond accordingly. Particularly for CAVs, it plays a crucial role in system malfunction detection, intelligent operations and cybersecurity protection. Growing efforts have been put into this area during the past years. For example,~\cite{alheeti2017using, ali2018intelligent} introduced intrusion detection methods for CAVs using deep learning and discriminant analysis. Additionally,~\cite{berlin2016poster} proposed a CAV misbehaviour detection method for service management. 

In this paper, we develop a deep autoencoder approach for anomaly detection in CAVs. As an unsupervised method, autoencoders are capable of finding the latent patterns of data. It is generally accepted that most of the research activities in this area are carried out using synthetic datasets, generated by means of simulation frameworks~\cite{rajbahadur2018survey}. Similarly, in our work, we will train and validate our deep autoencoder model using anomalous-free data, generated using OMNeT++ network simulator~\cite{omnetpp}. Later on, we introduce different abnormalities on our test dataset in order to evaluate the validity of our model.


The rest of this paper is organized as follows. Section~\ref{sec:method} presents our problem description and our deep autoencoder approach to detect anomalies.
Section~\ref{sec:simulator} gives more insights about the generation of our synthetic dataset, the tools used for that and the details about our scenario. Section~\ref{sec:results} presents the results and our analysis. Finally, Section~\ref{sec:conclusion} concludes the paper summarizing our findings.

\section{Methodology}
\label{sec:method}
\subsection{Problem Statement}
\label{sec:problem statement}
\begin{figure}[t]
    \centering
    \includegraphics[height = 4.0 cm]{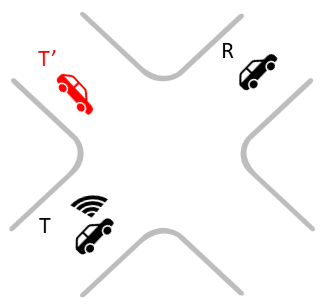}
    \caption{Self-reported location anomaly in CAVs. Here \eqr{T} is the transmitter in its real location, while \eqr{T^{\prime}} is ``ghost'' transmitter. \eqr{R}, being the receiver, receives the beacon from \eqr{T} with the faulty location of \eqr{T^{\prime}}.}
    \label{fig:problem_description}
\end{figure}

We consider a system where CAVs exchange beacons on a periodic basis. Also, we assume that anomalies are presented within the information encapsulated in the beacon frames, and more specifically at the self-reported CAV locations.
The falsified information is then received by other surrounding CAVs or infrastructure network. Fig.~\ref{fig:problem_description} gives a brief summary of the anomaly scenario considered in this paper. The fake position reported from the ``ghost'' transmitter \eqr{T^{\prime}} could be due to sensor malfunction in the actual transmitter \eqr{T}, or a possible information hacking during the transmission. The detection of self-reported location anomaly could further help with the root cause analysis and decision making in CAVs.

We aim to design a self-learning process for detecting the self-reported location anomalies. To do so, the Received Signal Strength Indicator (RSSI) is chosen as a proxy for the distance separation between two CAVs. The RSSI represents the beacon's signal strength and it is dependent on the maximum broadcasting power, the antenna gains, and the attenuation from the channel and the distance.
RSSI has been widely used for indoor localization~\cite{mcconville2018understanding,byrne2018residential, mcconville2019dataset},
human activity recognition~\cite{mukherjee2018rssi} and movement tracking~\cite{li2018indoor} in wireless networks. For CAVs, when a packet is received, the RSSI along with the transmitter-reported locations and the receiver self-location could form a strong state description for the self-reported location anomaly detection.

For our system, we will consider the RSSI between the different pairs of CAVs exchanging beacons in a Vehicle-to-Vehicle (V2V) fashion. As shown in Fig.~\ref{fig:problem_description}, \eqr{T} and \eqr{R} are the real transmitter-receiver pair, while \eqr{T^{\prime}} is a ``ghost'' transmitter. Locations of \eqr{T}, \eqr{R}, and \eqr{T^{\prime}} are represented as \eqr{l_{T}}, \eqr{l_{R}} and \eqr{l_{T^{\prime}}}, respectively. The distance between each pair of \eqr{T}, \eqr{R} and \eqr{T^{\prime}} are \eqr{D_{T,R}}, \eqr{D_{T,T^{\prime}}} and \eqr{D_{T^{\prime},R}}, respectively.

\subsection{The Deep Autoencoder Approach}
We design a Deep Autoencoder (DAE) for CAV self-reported location anomaly detection in an unsupervised manner. Autoencoders are a special type of artificial neural network, encoding high-dimensional data into a latent space by replicating the input in the output~\cite{kingma2013auto}. The idea for training a DAE for anomaly detection is to feed anomaly-free data into the network so it can learn the anomaly-free manifold and the corresponding latent space. Once the model has learned the anomaly-free manifold for a specific task, the error between the DAE input and output would be a strong indicator for recognizing anomalous samples. 

\begin{figure}[t]
    \centering
    \includegraphics[width = 3.0 in]{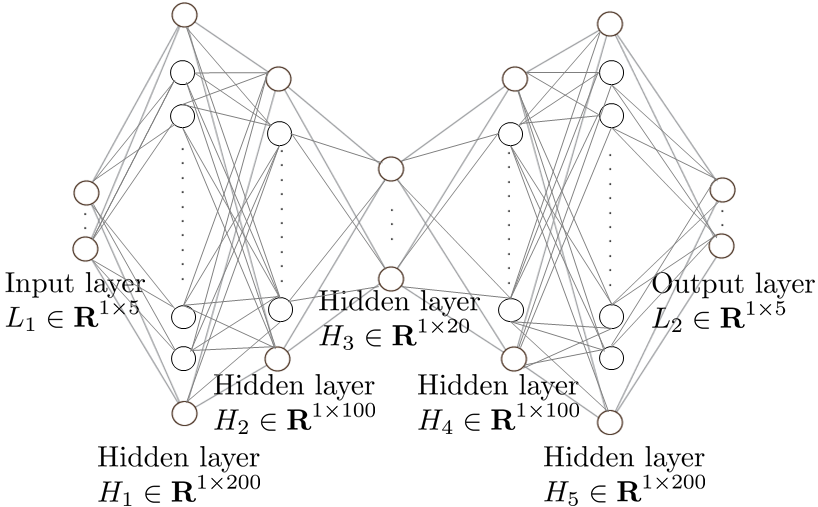}
    \caption{The seven-layer, fully-connected deep autoencoder structure designed in this paper.}
    \label{fig:AE_structure}
\end{figure}

Here we use a seven-layer, fully-connected autoencoder structure, as shown in Fig.~\ref{fig:AE_structure}. Note that the first hidden layer \eqr{H_1} has more neurons than the input layer \eqr{L_1} in our model. \eqr{L_1} can be seen as a data interpolation layer, which helps the model to learn the proper latent space. Symmetrically, the output layer \eqr{L_2} transfers the interpolated data into its original dimension. Hidden layers \eqr{H_1} to \eqr{H_5} build the standard DAE structure, with a latent space \eqr{L\in \mathds{R}^{1\times 20}}. The activation function is discarded at the bottleneck \eqr{H_3} and the output layer \eqr{L_2}. Dropout is not used in our model.

\subsubsection{Training}
As mentioned in Sec.~\ref{sec:problem statement}, the location of the receiver \eqr{l_R \in \mathds{R}^{1\times2}}, the self-reported transmitter location \eqr{l_T \in \mathds{R}^{1\times2}} (when there is no anomaly during transmission) or \eqr{l_{T^{\prime}} \in \mathds{R}^{1\times 2}} (when anomaly happens), along with the RSSI value \eqr{V_{\text{RSSI}}} form the feature set for self-reported location anomaly detection in CAVs. During training, anomaly-free samples \eqr{X = [l_R, V_{\text{RSSI}}, l_T]\in \mathds{R}^{1\times 5}} are fed to the DAE. The model is trained by minimizing the loss function \eqr{L_t}:
\begin{equation}
    \min_{\Theta} L_{t}(\Theta) = ||X - X^{\prime}||_{2}
\end{equation}
Here \eqr{X^{\prime}} is the output of the DAE, and \eqr{\Theta} represents the parameters. The goal is to train the DAE generating \eqr{X^{\prime}} close to \eqr{X}. Gradient descent optimization is used for training this model, with the learning rate of 0.00095.

\subsubsection{Validation}
To validate the performance of the trained DAE, the anomaly-free data is split into the training set and validation set with the proportion of 0.8 and 0.2, following the same data distribution. During validation, samples from the validation set are fed into the trained model. The validation loss is calculated as:
\begin{equation}
    L_{v} = ||X_{v} - X^{\prime}_{v}||_{2}
\end{equation}
Since DAE is an unsupervised method, here we introduce the adjusted mutual information (AMI) score~\cite{vinh2010mutualinformation} to evaluate the relation between \eqr{L_{t}} and \eqr{L_{v}}. It gives an unbiased evaluation of our trained model. The mean value and variance of \eqr{L_{t}} and \eqr{L_{v}} are also calculated. Results are shown in Table~\ref{tab:test loss}. Though the mean values and variances of two loss distributions are slightly different, they can be considered as nearly identical in the context of AMI. 
\begin{table}[htb]
    \centering
    \caption{Model evaluation by testing}
    \label{tab:test loss}
    \begin{tabular}{c|c|c|c}
        & AMI & Mean value & Variance\\
        \hline \hline
        Training loss & \multirow{2}*{1.0} & \eqr{4.11\times10^{-5}}& \eqr{8.13 \times 10^{-9}}\\
        Testing loss & & \eqr{5.37\times10^{-5}}& \eqr{5.36 \times 10^{-9}}\\
        \hline
    \end{tabular}
\end{table}

\subsubsection{Anomaly Detection}
After the DAE model is well-tuned and validated, it can be applied for anomaly detection. Potential anomalous samples \eqr{Y} go through the DAE, with an output \eqr{Y^{\prime}}. The difference between \eqr{Y} and \eqr{Y^{\prime}} can then be used for anomaly detection, while \eqr{L_{t}} and \eqr{L_{v}} serve as references.

\section{Data Generation}\label{sec:simulator}
For this particular work, we assume that each vehicle generates one beacon per second. Each beacon is encapsulated in a UDP packet with a total length \SI{140}{\byte}. Each UPD packet is broadcast in the network to the surrounding vehicles. We choose a \SI{2}{\kilo\meter}\eqr{\times} \SI{2}{\kilo\meter} area in central Bristol, UK as our simulation scenario. The number of vehicles within our system is constant (always $150$ vehicles). To simulate our scenario, we used OMNeT++~\cite{omnetpp} and our modified INET framework~\cite{parallelInet}. The vehicles mobility traces were generated using SUMO traffic generator~\cite{sumo} and parsed within our framework. Our INET framework was further modified with a logging interface that logs all the packets generated, transmitted and received, in a space-separated file format. These traces will be later used for our anomaly detection algorithm.

In particular, at the transmitter (TX) side, we find at first the wireless interface ID (e.g., \emph{ScenarioWorking.node[1].wlan[0].radio}), followed by the node ID (e.g., \emph{``1''} for the TX example). For this work, we assumed that all the vehicles are equipped with one IEEE 802.11p transceiver, operating at the frequency band of \SI{5.9}{\giga\hertz}. The next entry is the packet ID, i.e. \emph{``UDPData-50 1027''}, used to reconcile the transmitted with the received packets. \emph{UDPData-50} is the data structure called \emph{Signal} within INET, that represents the physical phenomena of transmitting a packet. The number following the signal is the sequence number of the event generated in INET. \emph{Start} fields represent the timestamp that the UPD packet started being transmitted (in seconds), followed by the position of the vehicle in space (given in meters). SUMO, when parses a real-world map, converts all the geolocation coordinates into a Cartesian plane with the southern-west map corner being the origin of the Euclidean space. Similarly, \emph{End} shows the timestamp that the transmission was over, followed by the position of the vehicle on that particular time.

The logging of each packet at the RX side follows a similar structure. Again we find the wireless interface ID followed by the node ID, the packet ID and the starting and ending timestamp of the reception as well as the positions of the RX vehicle. On top of that, our logging interface saves the RSSI of all the received packets (as it is being calculated within the INET framework). In order to calculate the RSSI for each packet, we take into account the building layout and the position of the vehicles. A scalar radio medium was chosen for our configuration, meaning that the analog signal power is represented with a scalar value over frequency and time. As a path loss model, we chose Rician Fading with a path loss exponent $\alpha=2.4$ and a Rician K-factor equal to $k=\SI{8}{\dB}$. Finally, the obstacle loss model chosen was the \emph{DielectricObstacleLoss}. This model calculates the dielectric and reflection loss along the straight path considering the shape, the position, the orientation, and the material of obstructing physical objects. The rest of our simulation parameters can be found in Table~\ref{tab:simParameters}. Also, the reasoning for choosing this setup was derived from the performance investigation in~\cite{agileCalibration}. 

A UDP packet is considered as deliverable under the following conditions. At first, the RSSI is compared with the sensitivity threshold \eqr{S_{\text{th}}} for the chosen Modulation and Coding Scheme (MCS). When the RSSI is lower than \eqr{S_{\text{th}}}, it is considered as non-deliverable. If the RSSI is above the \eqr{S_{\text{th}}}, then it is compared with the Signal-to-Noise-plus-Interference (SNIR) threshold \eqr{SNIR_{\text{th}}}. When below that, the packet is always considered as non-decodable due to errors introduced from the channel. The last case is when the RSSI is greater than the \eqr{SNIR_{\text{th}}}. For that, a Packet Error Rate (PER) value is calculated based on the current SNIR. This value is later compared with a random number chosen from a uniform random distribution and if it is greater, the packet is considered as delivered.
All the successfully received packets are logged in the dataset. These RX entries can then be reconciled with the transmitted ones using the combination of the values found at the packet ID (signal and sequence number), as mentioned above.

\begin{table}[t]
\renewcommand{\arraystretch}{1.07}
\centering
    \caption{Simulation Parameters.}
    \begin{tabular}{r|rl}

    \textbf{Parameter}           & \textbf{Value}    & \\ \hline \hline
    Simulation Time              & \SI{1800}         & \SI{}{\second}  \\
    Carrier Frequency            & \SI{5.9}          & \SI{}{\giga\hertz}  \\
    Bandwidth                    & \SI{10}           & \SI{}{\mega\hertz}  \\
    Path Loss Model              & Rician Fading     & \\
    Path Loss Exponent $\alpha$  & $2.4$             & \\
    Rician K-factor $k$          & \SI{8}            & \SI{}{\dBm} \\
    UDP Packet Length            & \SI{140}          & \SI{}{\byte}  \\
    UDP TX Interval              & \SI{1}            & \SI{}{\second}  \\
    TX Power                     & \SI{27}           & \SI{}{\dBm} \\
    TX/RX Antenna Gain           & \SI{9}            & \SI{}{\dBi} \\
    TX/RX Modulation             & QPSK              & \\
    RX Sensitivity \eqr{S_{\text{th}}}              & \SI{-88}          & \SI{}{\dBm} \\
    SNIR Threshold \eqr{SNIR_{\text{th}}}              & \SI{10}           & \SI{}{\dB} \\
    Background Noise             & $\mathcal{N}\left( -110, 3 \right)$ & \SI{}{\dBm} \\
    Connector and Cable Losses   & 3                 & \SI{}{\dBm} \\ \hline
    \end{tabular}
\label{tab:simParameters}
\end{table}


\section{Results and Discussions}
\label{sec:results}

To evaluate the proposed algorithm, an information hacking process has been imitated. More specifically, we consider a number of CAVs as hacked and force them to report random wrong locations. In this process, the inaccessible areas for CAVs, like buildings, grass-areas and rivers are avoided to increase the reasonableness of samples. Anomalous samples have the format of \eqr{[l_R, V_{RSSI}, l_{T^{\prime}}]}, while normal samples are \eqr{[l_R, V_{RSSI}, l_{T}]} without hacking. Here \eqr{l_{T^{\prime}}} is CAV-accessible locations in our simulation area. Anomaly detection is carried out for each received packet.

To conduct quantitative evaluation, anomalous samples are classified into 8 datasets by \eqr{D_{T,T^{\prime}}}. Table~\ref{tab:anomaly datasets} shows the details of each anomaly dataset, with the corresponding \eqr{D_{T,T^{\prime}}} range and mean values. Besides, One-class Support Vector Machine (SVM) method has been introduced as a benchmark to the proposed DAE method~\cite{chang2011libsvm}. One-class SVM is an unsupervised learning approach for novelty and outlier detection. By fitting a hyperplane to delineate the training points, new data can be classified as similar to the training data or not. Here we use the one-class SVM with a linear kernel.


\begin{table}[htb]
    \caption{Anomaly datasets (AD) (distance unit:\si{\meter})}
    \label{tab:anomaly datasets}
    \centering
    \begin{tabular}{c|c|c|c}
        &  Number of samples & \eqr{D_{T,T^{\prime}}}& \eqr{\overline{D_{T,T^{\prime}}}}\\
         \hline \hline
        AD1 & 1000 & [0,10) & 7.16\\
        AD2 & 1000 & [10,20) & 15.77\\
        AD3 & 1000 & [20,30) & 27.30\\
        AD4 & 1000 & [30,40) & 35.70\\
        AD5 & 1000 & [40,50) & 46.43\\
        AD6 & 1000 & [50,100) & 78.94\\
        AD7 & 1000 & [100,500) & 347.42\\
        AD8 & 1000 & \eqr{\geq 500} & 711.12\\
    \hline
    \end{tabular}
\end{table}

\subsection{Results}
\label{sec:AD results}
The performance of the anomaly detection model proposed in this paper is depicted with the Receiver Operating Characteristic (ROC) curve, along with the area under curve (AUC) value. ROC curve plots the correspondence between the False Positive Rate (FPR) and the True Positive Rate (TPR) over all possible classification thresholds. The data generated in Sec.~\ref{sec:simulator} serves as an anomaly-free baseline.

\begin{figure*}
    \centering
    \begin{subfigure}[c]{0.2\textwidth}
        \includegraphics[width = 1.7 in]{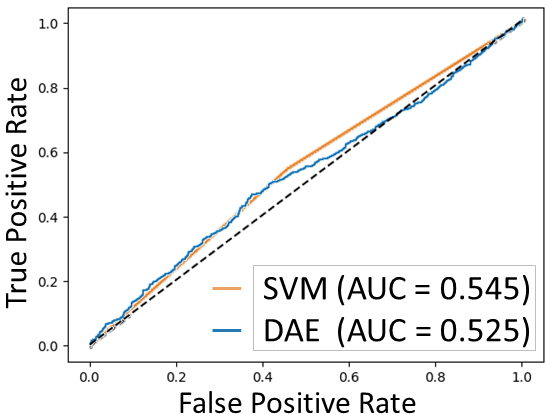}
        \caption{ROC curve for AD1.}
        \label{fig:1_10_roc}
    \end{subfigure}
    \hspace{0.6 cm}
    \begin{subfigure}[c]{0.2\textwidth}
        \includegraphics[width = 1.7 in]{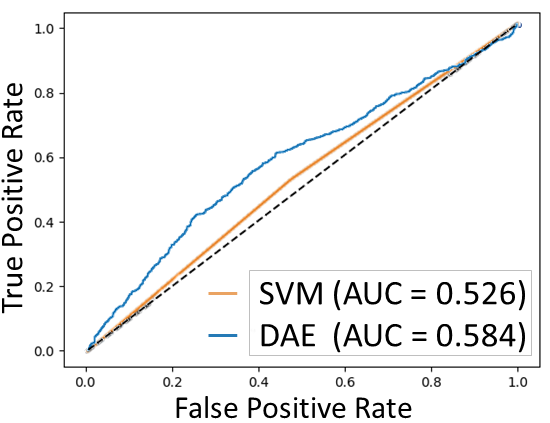}
        \caption{ROC curve for AD2.}
        \label{fig:10_20_roc}
    \end{subfigure}
    \hspace{0.6 cm}
    \begin{subfigure}[c]{0.2\textwidth}
        \includegraphics[width = 1.7 in]{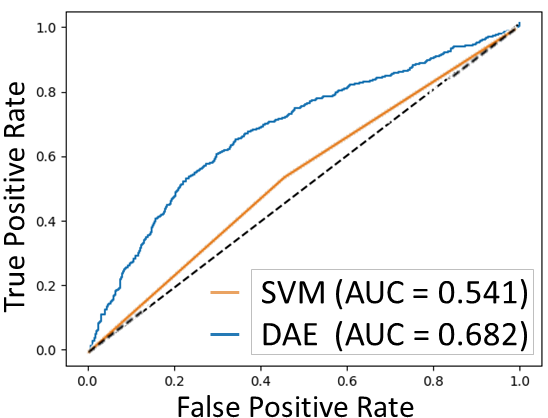}
        \caption{ROC curve for AD3.}
        \label{fig:20_30_roc}
    \end{subfigure}
    \hspace{0.6 cm}
    \begin{subfigure}[c]{0.2\textwidth}
        \includegraphics[width = 1.7 in]{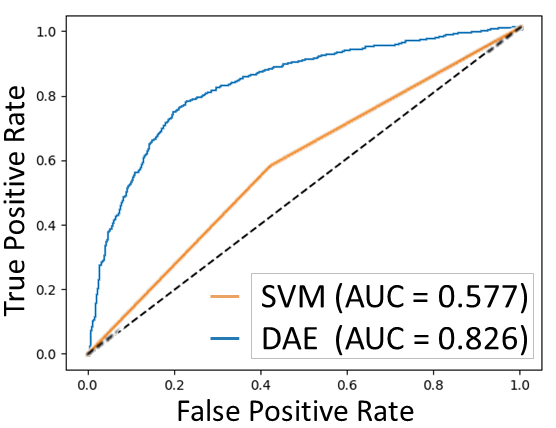}
        \caption{ROC curve for AD4.}
        \label{fig:30_40_roc}
    \end{subfigure} \\
    \begin{subfigure}[c]{0.2\textwidth}
        \includegraphics[width = 1.7 in]{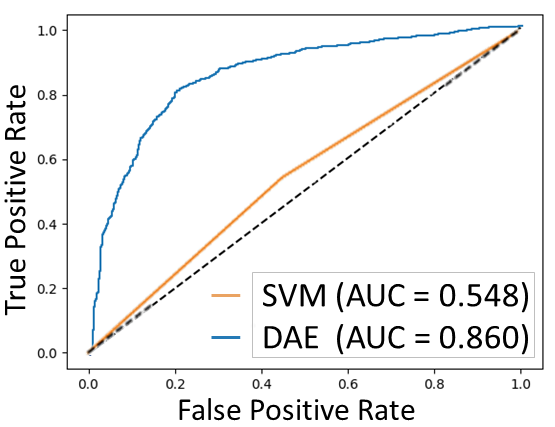}
        \caption{ROC curve for AD5.}
        \label{fig:40_50_roc}
    \end{subfigure}
    \hspace{0.6 cm}
    \begin{subfigure}[c]{0.2\textwidth}
        \includegraphics[width = 1.7 in]{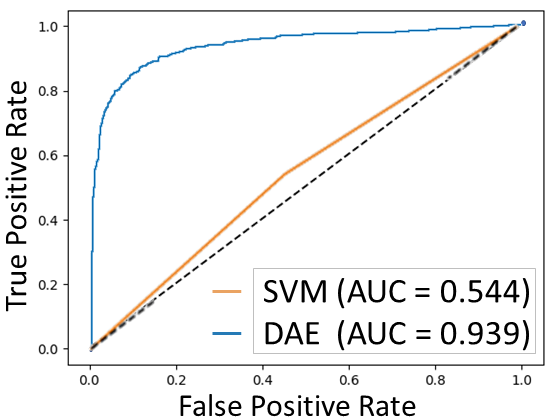}
        \caption{ROC curve for AD6.}
        \label{fig:50_100_roc}
    \end{subfigure}
    \hspace{0.6 cm}
    \begin{subfigure}[c]{0.2\textwidth}
        \includegraphics[width = 1.7 in]{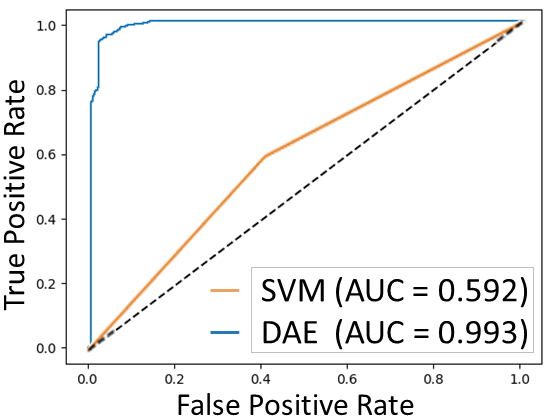}
        \caption{ROC curve for AD7.}
        \label{fig:100_500_roc}
    \end{subfigure}
    \hspace{0.6 cm}
    \begin{subfigure}[c]{0.2\textwidth}
        \includegraphics[width = 1.7 in]{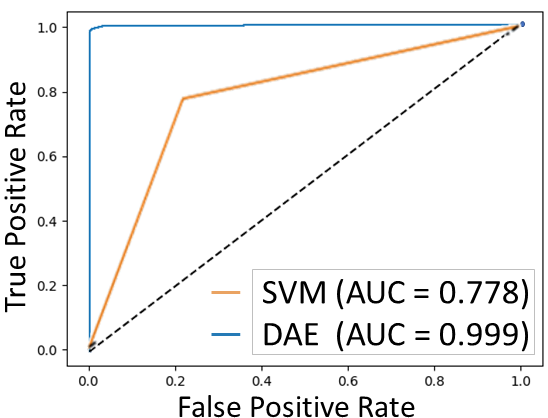}
        \caption{ROC curve for AD8.}
        \label{fig:500_roc}
    \end{subfigure}
    \caption{Anomaly detection results in AD1 to AD8.}
    \label{fig:ROC}
\end{figure*}

Fig.~\ref{fig:ROC} shows the detection results for AD1 to AD8, with AUC values in the right bottom of each subfigure. The diagonal black dash lines show where \eqr{\text{FPR} = \text{TPR}}. If the ROC curve falls along the diagonal line, the result is evaluated as no better than the random classification, where the AUC would be 0.5. From Fig.~\ref{fig:1_10_roc} we see the scenario where the faulty location \eqr{l_{T^{\prime}}} is close to the real location \eqr{l_T}, i.e., \eqr{D_{T,T^{\prime}} < \SI{10}{\meter}}. For that case, the chance to detect an anomaly almost equals to a random chance for both SVM and DAE. Under this situation, RSSI is not enough for self-reported location anomaly detection. When \eqr{D_{T,T^{\prime}}} reaches \SI{20}{\meter}, the detection rate is clearly boosted for DAE. When \eqr{D_{T,T^{\prime}}} is larger than \SI{30}{\meter}, there is a very good chance to detect anomalies with low FPR using DAE method. When the ghost vehicle \eqr{T^{\prime}} is \SI{100}{\meter} away from \eqr{T}, the detection rate for DAE is nearly 100\%, clearly better than SVM.

\begin{figure}[htbp]
    \centering
    \includegraphics[width = 2.2 in]{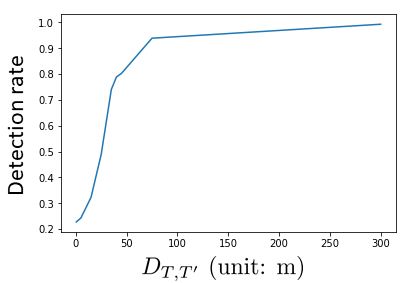}
    \caption{Anomaly detection rate under different distances \eqr{D_{T,T^{\prime}}}, with \eqr{\text{FPR} = 0.2}.}
    \label{fig:detection rate}
\end{figure}

To further investigate how the anomaly detection performance changes with \eqr{D_{T,T^{\prime}}}, we calculated the detection rate when \eqr{\text{FPR} = 0.2}. Fig.~\ref{fig:detection rate} shows the detection rate when \eqr{T^{\prime}} is located in the different radius ranges of \eqr{T}. We can also see that when \eqr{D_{T,T^{\prime}} < \SI{30}{\meter}}, it is difficult to detect self-reported location anomalies with the current model and features.

\subsection{Transmission Direction Anomaly Analysis}
According to the results in Sec.~\ref{sec:AD results}, when \eqr{D_{T,T^{\prime}} > \SI{30}{\meter}}, there is a very good chance to successfully detect abnormalities. In this section, we investigate another factor that could influence the detection performance. Consider a situation where \eqr{D_{T,T^{\prime}} > k}, here \eqr{k} is the detectable threshold. In the meantime, \eqr{|D_{T^{\prime},R} - D_{T,R}| < \epsilon}, where \eqr{\epsilon} is a small distance. Under such situation, the distance between the 'ghost' transmitter \eqr{T^{\prime}} and the real transmitter \eqr{T} is noticeable, while their distances to the receiver \eqr{R} are very close. It can be seen as a location anomaly, which mainly changes the data transmission direction with the same transmission distance. We would like to test the performance of our model in such a situation.

We generated two additional anomaly datasets, AD9 and AD10, to study the transmission direction anomaly. In AD9, \eqr{D_{T,T^{\prime}}> \SI{30}{\meter}} where \eqr{|D_{T,R} - D_{T^{\prime},R}| < \SI{1}{\meter}}. The wrong locations of \eqr{T^{\prime}} in AD9 imitate CAVs reporting wrong transmission directions. In AD10, we still have \eqr{D_{T,T^{\prime}}>\SI{30}{\meter}} while \eqr{|D_{T,R} - D_{T^{\prime},R}| \in [10,20] \si{\meter}}, for the sake of comparison. 

\begin{figure}[htb]
    \centering
    \begin{subfigure}[c]{0.2\textwidth}
        \includegraphics[width = 1.7 in]{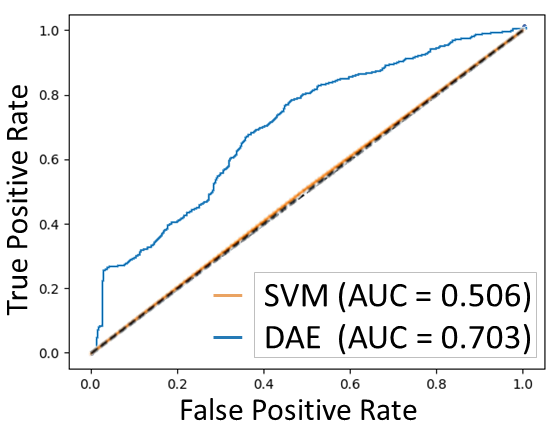}
        \caption{ROC curve for AD9.}
        \label{fig:AD9 ROC}
    \end{subfigure}
    \hspace{0.7 cm}
    \begin{subfigure}[c]{0.2\textwidth}
        \includegraphics[width = 1.7 in]{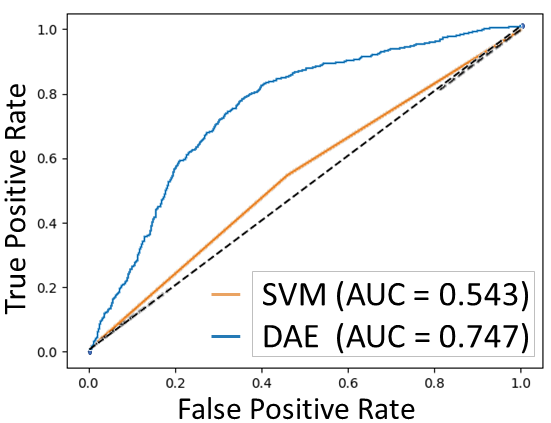}
        \caption{ROC curve for AD10.}
        \label{fig:AD10 ROC}
    \end{subfigure}
    \hspace{0.5 cm}
    \caption{Anomaly detection results in AD9 and AD10.}
    \label{fig:direction anomaly ROC}
\end{figure}

Detection results on AD9 and AD10 are shown in Fig.~\ref{fig:direction anomaly ROC}. In Fig.~\ref{fig:AD9 ROC}, even though \eqr{D_{T,T^{\prime}}} has reached 30\si{\metre}, the detection performance is not satisfying comparing to Fig.~\ref{fig:30_40_roc} to Fig.~\ref{fig:500_roc}. The detection rate of DAE is overall lower than the results in AD4 to AD8, with lower AUC value. The performance of SVM in AD9 is close to a random classification. In Fig.~\ref{fig:AD10 ROC}, the detection performance improved as \eqr{|D_{T,R} - D_{T^{\prime},R}|} increases. It can be seen as evidence that the trained DAE model mainly learns the correspondence between the transmitter/receiver distance \eqr{D_{T,R}} and the RSSI. Directional information is not learned because of the lack of relevant features. Data from multiple sources/sensors could be further considered for different types of anomalies.

\section{Conclusion}
\label{sec:conclusion}
In this study, we investigate the self-reported location anomaly detection problem in CAVs. To train a self-learning approach for anomaly detection, a deep autoencoder is designed. The proposed model is trained and tested on simulation data generated by OMNeT++, our modified INET framework and SUMO traffic generator, in an area of Bristol, UK.

Illustrative results about the self-reported location anomaly detection performance have been presented by only using the transmitter/receiver location pairs and RSSI values as features. We show that the DAE approach proposed in this paper is capable of detecting CAV location anomaly in a complicated scenario. We also present an insightful analysis of the detectable range of the proposed model, as well as the detection performance on transmission direction anomaly.

The proposed model connects unsupervised learning and anomaly detection in CAVs. It indicates that the powerful properties of unsupervised learning can bring benefits to future ITSs. The DAE approach proposed in this paper can be applied as a pre-processing step to the control centre of CAVs, verifying the clearness and reliability of the transmitted data before taking it to the CAV decision-making process.

\section*{Acknowledgment}
This work is partially funded by the Next-Generation Converged Digital Infrastructure (NG-CDI) Project, supported by British Telecommunications Group and Engineering and Physical Sciences Research Council (EPSRC), Grant ref. EP/R004935/1. It was also supported in part by Innovate UK Grant No. 102582 (Flourish Project).

\bibliographystyle{IEEEtran}
\bibliography{IEEEabrv,Ref}

\end{document}